# Acceleration of Convolutional Neural Network Using FFT-Based Split Convolutions

*Kamran Chitsaz[1], Mohsen Hajabdollahi[1], Nader Karimi[1], Shadrokh Samavi[1,2], Shahram Shirani[2]*
[1]Department of Electrical and Computer Engineering, Isfahan University of Technology, Isfahan, 84156-83111 Iran
[2]Department of Electrical and Computer Engineering, McMaster University, Hamilton, ON L8S 4L8, Canada

## ABSTRACT

Convolutional neural networks (CNNs) have a large number of variables and hence suffer from a complexity problem for their implementation. Different methods and techniques have developed to alleviate the problem of CNN's complexity, such as quantization, pruning, etc. Among the different simplification methods, computation in the Fourier domain is regarded as a new paradigm for the acceleration of CNNs. Recent studies on Fast Fourier Transform (FFT) based CNN aiming at simplifying the computations required for FFT. However, there is a lot of space for working on the reduction of the computational complexity of FFT. In this paper, a new method for CNN processing in the FFT domain is proposed, which is based on input splitting. There are problems in the computation of FFT using small kernels in situations such as CNN. Splitting can be considered as an effective solution for such issues aroused by small kernels. Using splitting redundancy, such as overlap-and-add, is reduced and, efficiency is increased. Hardware implementation of the proposed FFT method, as well as different analyses of the complexity, are performed to demonstrate the proper performance of the proposed method.

## 1. INTRODUCTION

Convolutional neural networks (CNN) are structures with a strong capability for feature extraction. We use CNNs in different fields of computer vision, including image classification, semantic segmentation, scene understanding, medical image analysis, etc. [1]–[3]. By using CNNs, a nonlinear model is trained to map an input space to a corresponding output space. This model has a large number of parameters that could cause problems in the implementation of this model. This problem is exacerbated in situations where there is a lack of available hardware resources [4].

Among a pool of parameters in CNNs, a lot of them are redundant [5]. A minor part of the network conducts a significant part of the computations of a CNN model. Since the emerging different CNN models, researchers have been working on simplifying their structure and reduce their redundancy. CNNs can be simplified from different perspectives. Quantization techniques are aiming to reduce the bit width representing the network parameters [6], [7]. Pruning techniques remove elements of the network which are not useful. Pruning is conducted on different levels, including connections, nodes, channels, and filters [5], [8]. Various studies are concentrating on designing a network which by itself has a simple structure. Neural architecture search (NAS) methods are methods that search for efficient structures manually or automatically [9], [10]. In some of the recent researches, the problem of design a simple structure is studied in the context of hardware implementation. In these methods, different techniques to have an efficient hardware implementation is under consideration [11]. Parallelization, data and resource sharing, and pipelining are some examples of hardware-based techniques to implement a suitable structure [12], [13].

Recently, the implementation of CNNs in the spectrum domain has been attracted substantial attention in the research community [14], [15]. CNN processing in the Fourier domain can be very beneficial. Some techniques have been proposed for the efficient implementation of FFT on hardware [16], [17]. There are two approaches in CNN processing in the FFT domain, which are implementing networks in the training phase and the inference phase. In the training and inference phases, FFT is used to reduce the number of computations of the CNN. In inference time, FFT can be used to result in an efficient hardware implementation [17], [18]. Most of the works in this area of research focused on replacing the convolution process by an element-wise production in the Fourier domain [14][19]. In [20], with respect to the benefits of linearity property of Fourier transform, the computational complexities for 3D filters are reduced. Also, in [21], a tile-based FFT is proposed to reduce the number of computations. However, in [20], [21], FFT implementation on dedicated hardware is not addressed directly. In [17], and [18], an FFT based on overlap and addition is proposed to reduce the memory access and computational complexity and implemented on dedicated hardware.

FFT improved the resource utilization and inference time of the CNN, but still, there are a lot of places for working on the FFT based inference of CNN. In FFT based computations, the problem of handling sizeable intermediate feature maps is less investigated. Also, the calculations of CNN in the FFT domain are still a lot. In this paper, a new FFT method for improving the CNN computations during inference is proposed in which CNN process patches of an input image. As a result, FFT is computed on image parts, and better memory management can be possible, and the number of computations is reduced. By modifying the size of the splitting, the number of redundant operations can be reduced, and the network can be processed according to the available resources. The rest of this paper is organized as follows. In Section 2, a brief review of FFT based CNN processing is described. In Section 3, the proposed method for FFT based CNN computing is demonstrated. Section 4 is dedicated to the experimental results, and finally, in Section 5, concluding points are presented.

## 2. CNN COMPUTING BASED ON FFT

Convolution is a significant part of the operations conducted by CNN. Suppose that we have a sample input $d \in \mathbb{R}^{N \times N}$ and a sample filter weight $w \in \mathbb{R}^{K \times K}$ with $k < N$. A single convolution can be represented as equation (1).

$$y = d \otimes w \qquad (1)$$

In which $\otimes$ is convolution operation. We can rewrite equation (1) in the Fourier domain as equation (2).

$$Y = FFT(d) \odot FFT(Padding(w)) = D \odot W \qquad (2)$$

where $\odot$ represents a *Hadamard* product and FFT represents the fast Fourier transform as equation (3).

$$D_{k,l} = \frac{1}{N^2} \sum_{n=0}^{N-1} \sum_{m=0}^{N-1} d_{m,n} e^{-\frac{2\pi i}{N^2}(nl+mk)} \qquad (3)$$

Padding is required to pad $w$ with some values, e.g., zero to make the size of $w$ equal to the size of $i$. In CNNs, $K \ll N$, hence, substantial padding is required. FFT can be calculated based on the Cooley-Tukey algorithm with the cost of $N^2 \log(N)$ and cost of *Hadamard* production is $N^2$. So the total cost of operations in equation (2) is $N^2 \log(N)$. Equation (2) has a redundancy due to using small size kernel, which is common in the CNN structures. This redundancy is addressed in references [16] to [18], and a new method to handle small size kernels are proposed. This method is named Overlap-and-Add convolution (OaAconv), as illustrated in Fig. 1. In this method, the input image is split into patches with the same size as the CNN kernel. FFT operations are conducted on these small patches, and after entry wise multiplications, they are transformed inversely to the spatial domain using an overlap addition. In Fig. 1, overlapped regions in output(y) are highlighted. As illustrated in Fig. 1 input image in this method should be split into patches with the same size of the filters, which are very small in the typical CNN structures. Also, redundancy is observed during overlapped addition, which conducted with a number comparable to the image's patches. Therefore, a large number of operations are required for overlapped additions.

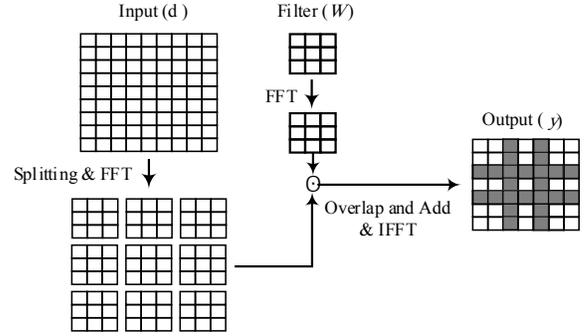

Fig. 1. Overlap and add method for FFT based convolution

## 3. FFT BASED COMPUTATION OF CNN USING SPLITTING APPROACH

### 3.1. Proposed method

In Fig. 2, the general flow of the proposed method is illustrated. A part of the CNN structure is considered in Fig. 2, which has three inputs as well as two output channels. In the first step, the input image is split into some patches. These patches can be of any size and can be selected based on the hardware requirements and other constraints. Suppose that a set $d$ including all of the input image splits $d_i$ as equation (4). The input image is reconstructed using the concatenation operation of $d_i \in d$. $d_i$ is a sample patch with the size of $S \times$ $S$, and a filter kernel $W$ has the size of $K \times K$. Since convolutional operations require some padding $p_i$, $d_i$ should be padded by $p_i$ and a patch set $\tilde{d}$ with sample patch $\tilde{d}_i$ is created as equation (5, 6). Therefore, the final patch $\tilde{d}_i$ have the size of $(T + K - 1) \times (T + K - 1)$. In other words, as illustrated in Fig. 2, patches are extracted from the input image by overlapping with the size of $\left\lfloor \frac{K}{2} \right\rfloor$. In this way, the convolutional operations in Equation (1), can be rewritten as the convolution of patches as Equation (7).

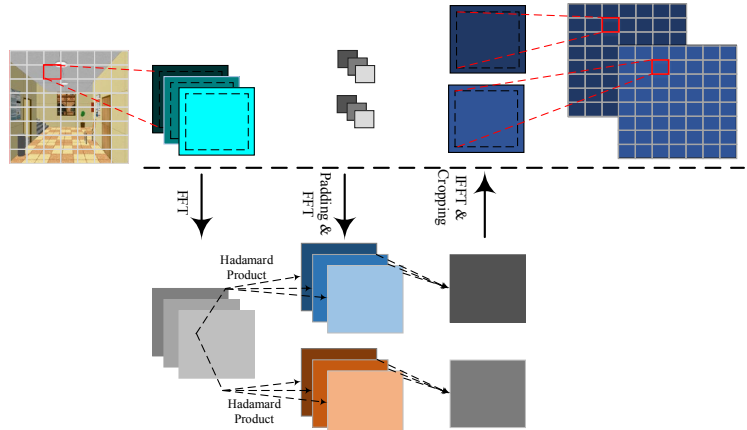

Fig. 2. The proposed procedure for the FFT-based processing of CNN using splitting.

$$d = \{d_1, d_2, \ldots, d_{|d|}\} \quad (4)$$

$$\check{d} = \{(d_1 + p_1), (d_2 + p_2), \ldots, (d_{|d|} + p_{|d|})\} \quad (5)$$

$$\tilde{d} = \{\tilde{d}_1, \tilde{d}_2, \ldots, \tilde{d}_{|d|}\} \quad (6)$$

$$y = \tilde{d} \otimes W = \{\tilde{d}_1 \otimes w, \tilde{d}_2 \otimes w, \ldots, \tilde{d}_{|d|} \otimes w\} \quad (7)$$

Although there is some redundancy due to overlapping operations, this redundancy with the small size kernels in CNNs, also selecting a relatively large patch size, can be negligible. According to Fig. 2, in each step, a patch with its padding is extracted from the input image and then is transformed into the FFT domain. Also, a convolutional filter $w$ should be transformed into the FFT domain. Due to the different size of the patches and filters, a zero-padding $p_0$ as equation (8) is applied to the filter before FFT to yield the same size of the extracted patches.

$$\widetilde{w} = w \ \& \ p_0 \quad (8)$$

In the FFT domain, a Hadamard product provides the result of convolution. A single Hadamard product in the FFT domain is illustrated in equation (9).

$$Y_i = FFT(\tilde{\iota}_i) \odot FFT(\widetilde{w}) = \tilde{I}_i \odot \widetilde{W} \quad (9)$$

The Hadamard production is an entry-wise operation and can be conducted with lower complexity than a convolution operation. All of the patches are processed based on the equation (9), and inverse FFT is conducted on them. After the IFFT transform, the resulted patch is larger than its actual size. Hence we need to crop the patch to make it suitable for concatenation. The set $y$ is calculated by Equation (10).

$$y = \{crop(\tilde{I}_1 \odot \widetilde{W}), crop(\tilde{I}_2 \odot \widetilde{W}), \ldots, crop(\tilde{I}_{|d|} \odot \widetilde{W})\} \quad (10)$$

The result of convolution is provided through a simple concatenating on the members of $y$. The major problem in the implementation of CNN on hardware is handling feature maps. There are lots of feature maps in the CNNs, and this problem is aggravated when the input is relatively large. The issue of handling the large feature maps is not addressed properly in FFT based methods. This problem is addressed in the proposed approach by using a splitting method and simultaneously reducing the number of redundant operations. In the proposed splitting approach, different splitting sizes can be employed, and thereby better handling of the large feature maps is possible. Also, the splitting method can be employed in case of other filters, which reduce the number of memory access to read input data.

### 3.2. Computational complexity and memory access analysis

If we consider the equation (1), convolutional operations in the spatial domain need $(N - k + 1)^2 K^2$ multiplications and about $(N - k + 1)^2 K^2$ additions. Also, an element of input data (with stride 1) is accessed K times, and total memory requirement is $N^2 + K^2$ times. Conventional methods for FFT based CNN computing have a cost of FFT computing equal to the $N^2 \log N$ times for the number of multiplications and additions, and a cost of multiplication as $N^2$ for Hadamard production. So a total cost of $N^2 \log N$ could be considered for CNN processing using a conventional FFT method. Also, in this method, the memory storage requirement is $2N^2$ and each location is accessed one time. In [16]–[18], overlap and add approach is proposed in which processes images in its constituent blocks with a size equal to the filter kernels. It is stated that the total complexity of their method is $N^2 \log k$, mthe emory storage requirement is $N^2 + K^2$ and each memory location containing input data is accessed one time.

In the proposed method, an image patch with a size of S is assumed. FFT of any patches have a cost equal to $O(S^2 \log S^2)$ and there are $\frac{N^2}{S^2}$ patches in an image, so the total cost is $N^2 \log S$. The size of the memory requirement is $N^2 + S^2$ and each location of memory containing data is accessed only once. If we consider the size of the patches equal to the size of kernels, a similar complexity with respect to the OaAConv method can be obtained. However, there are some notes about the comparison of the proposed method with the OaAConv method. The first one is the problem of the number of computational operations. By looking at the OaAConv method, we can correctly write the total number of operations. The number of multiplications in OaAConv method can be computed based on equation (11) which named as $\#MUL_{\text{OaAConv}}$.

$$\#MUL_{\text{OaAConv}} = \frac{N^2}{k^2} ( 2(2k - 1)^2 \log(2k - 1)^2 + (2k - 1)^2 ) \quad (11)$$

Which are due to the multiplication operations in FFT using the Cooley-Tukey algorithm, Hadamard production, and inverse FFT. There are additional operations in FFT using the Cooley-Tukey algorithm, inverse FFT, and overlap of additions. The number of additions ($\#ADD_{\text{OaAConv}}$) can be computed as equation (12).

$$\#ADD_{\text{OaAConv}} = \frac{N^2}{k^2} ( 2(2k - 1)^2 \log(2k - 1)^2 + k^2 - k ) \quad (12)$$

We can perform similarly for the proposed method to count the number of multiplications and additions, which we name them as $\#MUL_{\text{SplitConv}}$ and $\#ADD_{\text{SplitConv}}$, represented in equations (13) and (14), respectively. All of the operations in equations (13) and (14) are due to the Cooley-Tukey operations for FFT and inverse FFT.

$$\#MUL_{\text{SplitConv}} = \frac{N^2}{s^2}(2(s+k-1)^2 \log(s+k-1)^2 + (s+k-1)^2) \quad (13)$$

$$\#ADD_{\text{SplitConv}} = \frac{N^2}{s^2}(2(s+k-1)^2 \log(s+k-1)^2) \quad (14)$$

Based on equations (11) to (14) it is possible to visualize the total number of operations based on different parameters.

In Fig. 3, the number of computational operations in two methods OaAConv as well as the proposed names as SplitConv, are illustrated for S=16 and S= 32. It can be observed that the number of total operations in the proposed method is less than the OaAConv method. By increasing the kernel size, more differences can be observed. Figure 3, shows that the computational operations can be reduced efficiently by the proposed method. From the memory analysis perspective, it is worthwhile to say that by employing relatively small size patches, small size feature maps are generated, and memory management problems can be better handled. In the conventional FFT method, all the input data elements should be accessed simultaneously, which causes a problem, especially in the case of the large input image. Thanks to using the patches of input data, convolutional operations can be conducted with more independence. Also, a patch can be fetched only once, and better efficiency is achieved.

## 4. EXPERIMENTAL RESULTS

To evaluate the proposed method on a practical experiment, we implement the proposed method on hardware. Proposed FFT based convolution is described using Verilog hardware description language. A Xilinx FPGA XC6VLX240T is employed as a target device using Xilinx Vivado tools. All of the implemented codes on the FPGA are evaluated by its software counterpart using Python programming language.

For testing, an 8×8 patch is considered for our proposed method. At first, an FFT based convolution for 8×8 block with a single channel as input is designed. The synthesis report from our implementation is illustrated in Table 1. Also, a convolution method in the spatial domain is implemented, and related synthesis results are reported in Table 1. In single-channel input, the proposed method has not any improvement. For a better comparison of the performance in the methods mentioned above, spatial convolution is designed in such a way that both of the methods consume the same resources. Based on Table 1, convolutional layers on the VGG16 network are implemented, and their run times are analyzed and compared in Fig. 4. It can be observed that the run time of our proposed FFT method is shorter than the convolution in the spatial domain. Although in the single-channel input, as illustrated in Table. 1, the proposed method has not any improvement, but by increasing the number of filter channels, a significant improvement can be observed for the proposed method.

## 5. CONCLUSION

An efficient method for FFT based CNN processing was proposed. The proposed method split input image to patches, and all of the FFT related operations were conducted on them. We showed that by using the splitting approach, the number of multiplications and additions were reduced as compared to state of the art OaAConv. Hardware simulations demonstrated that an appropriate implementation resulted. For the convolutional layers in the VGG16 network, a significant improvement was observed with comparison to the spatial convolution. Finally, it can be said that the splitting approach can be very beneficial for reducing the cycles and operations required to process CNNs and prevents multiple access to the same data.

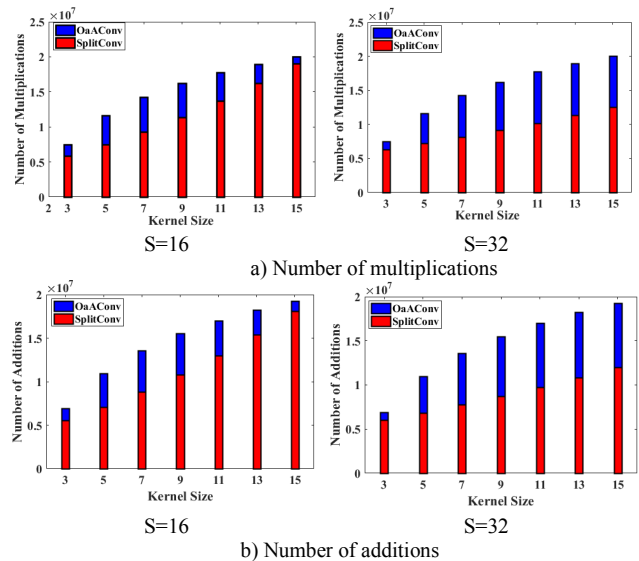

Fig. 3. Number of operations in OaAConv and SplitConv.

Table 1. Complexity of the spatial and proposed convolution with a single channel as input.

|  | # of LUT | # of Registers | # of DSP | Latency (cycle) |
|---|---|---|---|---|
| Spatial Conv. | 63531 | 74079 | 612 | 104 |
| SplitConv. | 49200 | 58708 | 256 | 395 |

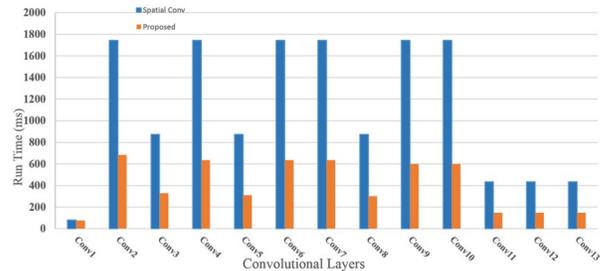

Fig. 4. Run time of implemented methods for computing convolution in VGG layers.

## REFERENCES

[1] Y. Lecun, Y. Bengio, and G. Hinton, "Deep learning," *Nature*, vol. 521(7553), pp. 436–444, 2015.

[2] E. Shelhamer, J. Long, and T. Darrell, "Fully Convolutional Networks for Semantic Segmentation," *IEEE conference on computer vision and pattern recognition (CVPR), pp. 3431-3440, 2015.*

[3] M. Hajabdollahi, R. Esfandiarpoor, E. Sabeti, N. Karimi, S. M. R. Soroushmehr, and S. Samavi, "Multiple abnormality detection for automatic medical image diagnosis using bifurcated convolutional neural network," *Biomed. Signal Process. Control*, vol. 57, pp. 101792, 2020.

[4] V. Sze, Y.-H. Chen, T.-J. Yang, and J. S. Emer, "Efficient processing of deep neural networks: A tutorial and survey," *Proc. IEEE*, vol. 105(12), pp. 2295–2329, 2017.

[5] Y. Guo, A. Yao, and Y. Chen, "Dynamic network surgery for efficient DNNs," in *Advances in Neural Information Processing Systems*, pp. 1379-1387, 2016.

[6] D. D. Lin, S. S. Talathi, and V. S. Annapureddy, "Fixed Point Quantization of Deep Convolutional Networks," *International Conference on Machine Learning*, pp. 2849-2858, 2016.

[7] J. Wu, C. Leng, Y. Wang, Q. Hu, and J. Cheng, "Quantized convolutional neural networks for mobile devices," *IEEE Conference on Computer Vision and Pattern Recognition (CVPR)*, pp. 4820-4828, 2016.

[8] M. M. Pasandi, M. Hajabdollahi, N. Karimi, and S. Samavi, "Modeling of Pruning Techniques for Deep Neural Networks Simplification," *arXiv Prepr. arXiv2001.04062*, 2020.

[9] M. Wistuba, A. Rawat, and T. Pedapati, "A survey on neural architecture search," *arXiv Prepr. arXiv1905.01392*, 2019.

[10] E. Malekhosseini, M. Hajabdollahi, N. Karimi, and S. Samavi, "Modeling Neural Architecture Search Methods for Deep Networks," *arXiv Prepr. arXiv1912.13183*, 2019.

[11] A. Ardakani, C. Condo, M. Ahmadi, and W. J. Gross, "An Architecture to Accelerate Convolution in Deep Neural Networks," *IEEE Trans. Circuits Syst. I Regul. Pap.*, vol. 65(4), pp.1349-1362, 2018.

[12] Y. H. Chen, J. Emer, and V. Sze, "Using Dataflow to Optimize Energy Efficiency of Deep Neural Network Accelerators," *IEEE Micro*, vol. 37(3), pp.12-21, 2017.

[13] Y. Wang, J. Lin, and Z. Wang, "An Energy-Efficient Architecture for Binary Weight Convolutional Neural Networks," *IEEE Transactions on Very Large Scale Integration (VLSI) Systems*, vol. 26(2), pp.280-293, 2017.

[14] M. Mathieu, M. Henaff, and Y. LeCun, "Fast training of convolutional networks through FFTS," *International Conference on Learning Representations (ICLR),* 2014.

[15] O. Rippel, J. Snoek, and R. P. Adams, "Spectral representations for convolutional neural networks," in *Advances in neural information processing systems*, pp. 2449-2457, 2015.

[16] T. Abtahi, A. Kulkarni, and T. Mohsenin, "Accelerating convolutional neural network with FFT on tiny cores," *IEEE International Symposium on Circuits and Systems (ISCAS)*, pp. 1-4, 2017.

[17] T. Abtahi, C. Shea, A. Kulkarni, and T. Mohsenin, "Accelerating Convolutional Neural Network with FFT on Embedded Hardware," *IEEE Trans. Very Large Scale Integr. Syst.*, vol. 26(9), pp.1737-1749, 2018.

[18] T. Highlander and A. Rodriguez, "Very Efficient Training of Convolutional Neural Networks using Fast Fourier Transform and Overlap-and-Add," *arXiv preprint arXiv:1601.06815*, 2015.

[19] A. Lavin and S. Gray, "Fast algorithms for convolutional neural networks," in *Proceedings of the IEEE Conference on Computer Vision and Pattern Recognition*, pp. 4013-4021, 2016.

[20] N. Nguyen-Thanh, H. Le-Duc, D.-T. Ta, and V.-T. Nguyen, "Energy efficient techniques using FFT for deep convolutional neural networks," *International Conference on Advanced Technologies for Communications (ATC)*, pp. 231–236, 2016

[21] J. Lin and Y. Yao, "A Fast Algorithm for Convolutional Neural Networks Using Tile-based Fast Fourier Transforms," *Neural Process. Lett.*, vol. 50(2), pp. 1951–1967, 2019.